\documentclass[10pt,twocolumn,letterpaper]{article}

\usepackage{iccv}
\usepackage{times}
\usepackage{epsfig}
\usepackage{graphicx}
\usepackage{amsmath}
\usepackage{amssymb}

\usepackage{bm}
\usepackage{tabularx}
\usepackage{makecell}

\usepackage{algorithm, algorithmic}

\usepackage[pagebackref=true,breaklinks=true,letterpaper=true,colorlinks,bookmarks=false]{hyperref}

\ificcvfinal\fi

\begin{document}

\title{IC-FPS: Instance-Centroid Faster Point Sampling Module for 3D Point-base Object Detection}

\author{Haotian Hu\\
Leapmotor\\
{\tt\small hht1996ok@zju.edu.cn}
\and
Fanyi Wang\\
Zhejiang university\\
{\tt\small 11730038@zju.edu.cn}
\and
Jingwen Su\\
 OPPO Research Institute\\
{\tt\small j\_su95@outlook.com}
\and
Shiyu Gao\\
UCAS\\
{\tt\small gaoshiyu@ict.ac.cn}
\and
Zhiwang Zhang\\
The University of Sydney\\
{\tt\small zhiwang.zhang@sydney.edu.au}
}

\maketitle
\ificcvfinal\thispagestyle{empty}\fi

\begin{abstract}

3D object detection is one of the most important tasks in autonomous driving and robotics. 
Our research focuses on tackling low efficiency issue of point-based methods on large-scale point clouds. 
Existing point-based methods adopt farthest point sampling (FPS) strategy for downsampling, which is computationally expensive in terms of inference time and memory consumption when the number of point cloud increases. 
In order to improve efficiency, we propose a novel Instance-Centroid Faster Point Sampling Module (IC-FPS) , which effectively replaces the first Set Abstraction (SA) layer that is extremely tedious. 
IC-FPS module is comprised of two methods, local feature diffusion based background point filter (LFDBF) and Centroid-Instance Sampling Strategy (CISS). 
LFDBF is constructed to exclude most invalid background points, while CISS substitutes FPS strategy by fast sampling centroids and instance points.
IC-FPS module can be inserted to almost every point-based models.
Extensive experiments on multiple public benchmarks have demonstrated the superiority of IC-FPS.
On Waymo dataset \cite{sun2020scalability} , the proposed module significantly improves performance of baseline model and accelerates inference speed by 3.8 times. For the first time, \textbf{real-time detection} of point-based models in large-scale point cloud scenario is realized. Code to reproduce our results is available at \url{https://github.com/hht1996ok/IC-FPS}.\\
\end{abstract}

\section{Introduction}

In recent years, with rapid development of sensors such as LiDAR and millimeter wave radar, point cloud has been widely applied in 3D tasks as a common 3D representation. And 3D object detection plays a crucial role in autonomous driving. 
However, due to the orderless, sparse and irregular nature of point cloud, it is still challenging to predict 3D detection box with multiple degrees-of-free. 
Early works project point clouds to multi-view \cite{ali2018yolo3d, ku2018joint, lang2019pointpillars, yang2018pixor, zhou2020end, simon2019complexer, simony2018complex, lu2019scanet}, or voxelise point clouds \cite{zhou2018voxelnet, yan2018second, yi2020segvoxelnet, zhao2020sess, zheng2021cia, ye2020hvnet, deng2021voxel, liang2019multi} and extract features through 3D convolution. 
Although these methods have achieved outstanding results, it is unavoidable to lose information when point cloud is converted to intermediate representation such as block, resulting in degraded model performance. 

\begin{figure}
    \centering
    \includegraphics[width=0.47\textwidth, height=0.21\textheight]{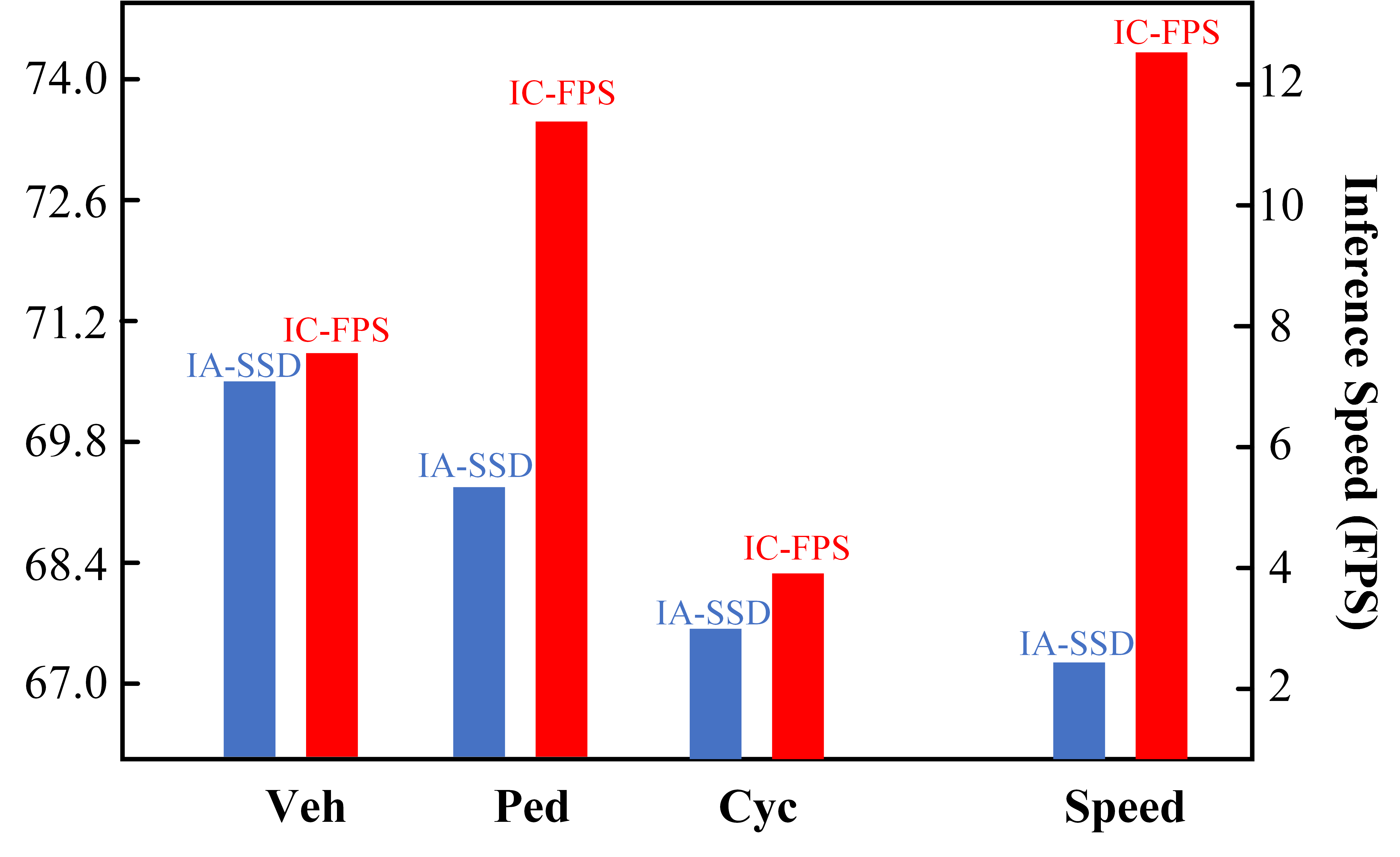}
    \caption{Comparison of performance and inference time of various models on Waymo dataset \cite{sun2020scalability}. Experiment results are derived by using OpenPCDet \cite{openpcdet2020} framework on a single A40 GPU. More details can be found in Table \ref{tab:tab2}.} 
    \label{fig: fig1}
\end{figure}

As we know, point-based methods \cite{qi2017pointnet,qi2017pointnet++,yang20203dssd,zhang2022not,qian2020end,thakur2020dynamic,wang2019dynamic,shi2020point} extract point cloud features layer by layer, thus rely on sophisticated downsampling strategies, such as Distance-Farthest Point Sampling (D-FPS) and Feature-Farthest Point Sampling (F-FPS) \cite{qi2017pointnet++,yang20203dssd,zhang2022not}, to obtain center points. 
But computational costs of these strategies are too expensive to afford when applied for large-scale 3D object detection. As shown in Table \ref{tab:tab1}, the first layer of downsampling consumes most of the inference time. 
On large datasets like Waymo \cite{sun2020scalability} and nuScenes \cite{caesar2020nuScenes}, when the number of input points reaches 100k, it takes 498.1ms for D-FPS to sample 16,384 points, which severely hinders the applicability of point-based models on tackling large-scale point cloud tasks. 
While for 3D object detection, to the best of our knowledge, existing works have not provided a sampling strategy that attends efficiency and effectiveness simultaneously.


Moreover, as an object-oriented task, 3D detection does not require overly dense representations of background context.
SASA \cite{chen2022sasa} employs MLP to encode point features, combing with FPS to increase the number of instance points. 
While IA-SSD \cite{zhang2022not} utilizes MLP to replace the last two layers of FPS for center point selection, in order to improve the recall rates of instance targets.
Nevertheless, current methods depend on complicated downsampling strategies or the preceding SA layers to extract neighboring features, for differentiating foreground and background points. And such operation includes a large amount of computations in the first a few SA layers that are inefficient.


Firstly, to improve speed of downsampling module, we propose Centroid Instance Sampling Strategy (CISS) as an alternative of FPS. CISS significantly increases inference speed of the model and realizes real-time detection in large-scale point cloud scenes for point-based models.
In addition, we design a Centroid Point Offset Module to restore the raw geometric structure of instance targets, which helps the model to accurately regress instance size.
Besides, LFDBF effectively discriminates foreground and background blocks to reduce computation cost wasted in invalid background regions.
To further improve performance of LFDBF,  we propose Density-Distance Focal Loss to ensure sparse foreground points at a distance are effectively sampled.
Finally, we couple CISS and LFDBF and propose IC-FPS module. It solves the low efficiency problem of point-based models in large-scale point cloud scene, and significantly enhances performances of baseline.

In Section 4, we testify IC-FPS on multiple large-scale benchmarks and prove that the proposed module effectively alleviates inefficiency issue of the first SA layer.
As shown in Figure \ref{fig: fig1}, IC-FPS improves baseline performance and inference speed by a large margin on Waymo dataset, and demonstrates its effectiveness and efficiency.

To summarize, our contributions are listed as follows,
\begin{itemize}
  \item We introduce a point-based 3D detection module IC-FPS, which achieves efficient and accurate detection in large-scale point cloud scenes when inserted to existing point-based methods.
  \item We propose a Local Feature Diffusion Background Filter (LFDBF) to exclude invalid background points in the raw point cloud input and reduce inefficient computation.
  \item We propose Centroid Instance Sampling Strategy (CISS) to realize real-time inference of point-based models in large-scale point cloud scenarios, as an efficient alternative of complicated downsampling strategies such as FPS.
  \item Extensive experiments on multiple large datasets have demonstrated effectiveness and superiority of IC-FPS.
\end{itemize}

\section{Related Works}
Due to the intricate properties of point clouds, researchers attempt to project point clouds to multi-view or regular voxel grids to represent features \cite{zhou2018voxelnet,yan2018second,lang2019pointpillars,yi2020segvoxelnet,zhao2020sess,zheng2021cia,ye2020hvnet,deng2021voxel}. But these static projection methods result in information loss.
Point-based methods directly use raw point cloud information as the input, and aggregate global features of point cloud from the top to the bottom.
Section \ref{sec:Multi-view},  \ref{sec:voxel} and \ref{sec:point} introduce 3D object detection methods based on multi-view, voxel and point respectively.

\subsection{Multi-view Based Methods}
\label{sec:Multi-view}
Early works project unstructured point clouds to multiple 2D views, for the convenience of direct use of convolution operations.
MVCNN \cite{su2015multi} converts multiple views to global features via max pooling layers, however it inevitably leads to massive information loss. 
MV3D-Net \cite{chen2020multi} only uses top view and front view for feature extraction, which preserves primary feature information and reduces computation cost.

\subsection{Voxel-based Methods}
\label{sec:voxel}
In order to process unstructured 3D point cloud, voxel-based 3D detectors convert point clouds to regular voxel grids, such that the commonly used convolutions can be applied. 
VoxelNet \cite{zhou2018voxelnet} voxelizes the point cloud, and employs a block feature encoding layer to aggregate global and local information. However, computation and storage cost of 3D convolution increase along with resolution and bring unaffordable burden. 
SECOND \cite{yan2018second} alleviates this issue by introducing 3D sparse convolution to substitute traditional convolution, and effectively optimizes memory usage and computation speed. 
PointPillars \cite{lang2019pointpillars} further improves detection speed. It simplifies voxel to pillar with two dimensions, projects features to bird's eye view and applies 2D convolutions to extract deep features. 
SA-SSD \cite{he2020structure} employs segmentation and center point prediction to facilitate model for further extraction of structural information. 

\subsection{Point-based Methods}
\label{sec:point}
Different to voxel-based methods, point-based methods use original information as the input, and adopt top-down learning to extract unstructured features of point cloud. Existing point-based methods normally adopt architectures similar to PointNet++ \cite{qi2017pointnet++}, which aggregates features by using symmetric aggregation function.
PointRCNN \cite{shi2019pointrcnn} is the first 3D object detection model based on the original point cloud. It uses foreground segmentation network to obtain valid points for detection box regression, and predicts detection box by basing on bin. 
3DSSD \cite{yang20203dssd} is a single-stage detection framework that combines advantages of D-FPS and F-FPS.
IA-SSD \cite{zhang2022not} uses FPS and instance-aware downsampling modules to extract features point by point, and utilizes contextual clues around bounding box to predict centroids.
Nevertheless, existing methods cannot fully achieve fast and accurate downsampling. Constrained by complicated strategy in SA layer, existing methods are infeasible for large-scale 3D object detection.

The proposed IC-FPS module differs from the aforementioned methods. It circumvents the disadvantage that SASA and IS-SSD are unable to sample in the first SA layer, and directly optimizes the first downsampling that is the most time-consuming, thus substantially improves baseline efficiency.

\section{The Proposed Instance-Centroid Feature Diffusion Sampling Module}
\subsection{Overview}
Compared to dense perception tasks like 3D semantic segmentation, 3D object detection pays more attentions to target objects. Existing 3D detection models use a great amount of background points as the input, and accounts for massive inefficient computation in subsequent models.
Besides, complicated downsampling strategies such as FPS are unavoidable for existing point-based methods, causing  computation cost and inference time to increase. 
To address these problems, we propose the Instance-Centroid Feature Diffusion Sampling (IC-FPS) module that is applicable for point-based 3D object detection models. 
IC-FPS module is comprised of Local Feature Diffusion based Background Point Filter (LFDBF), Centroid-Instance Sampling Strategy (CISS) and SA layer.

\begin{figure*}[th!]
    \centering
    \includegraphics[scale=0.7]{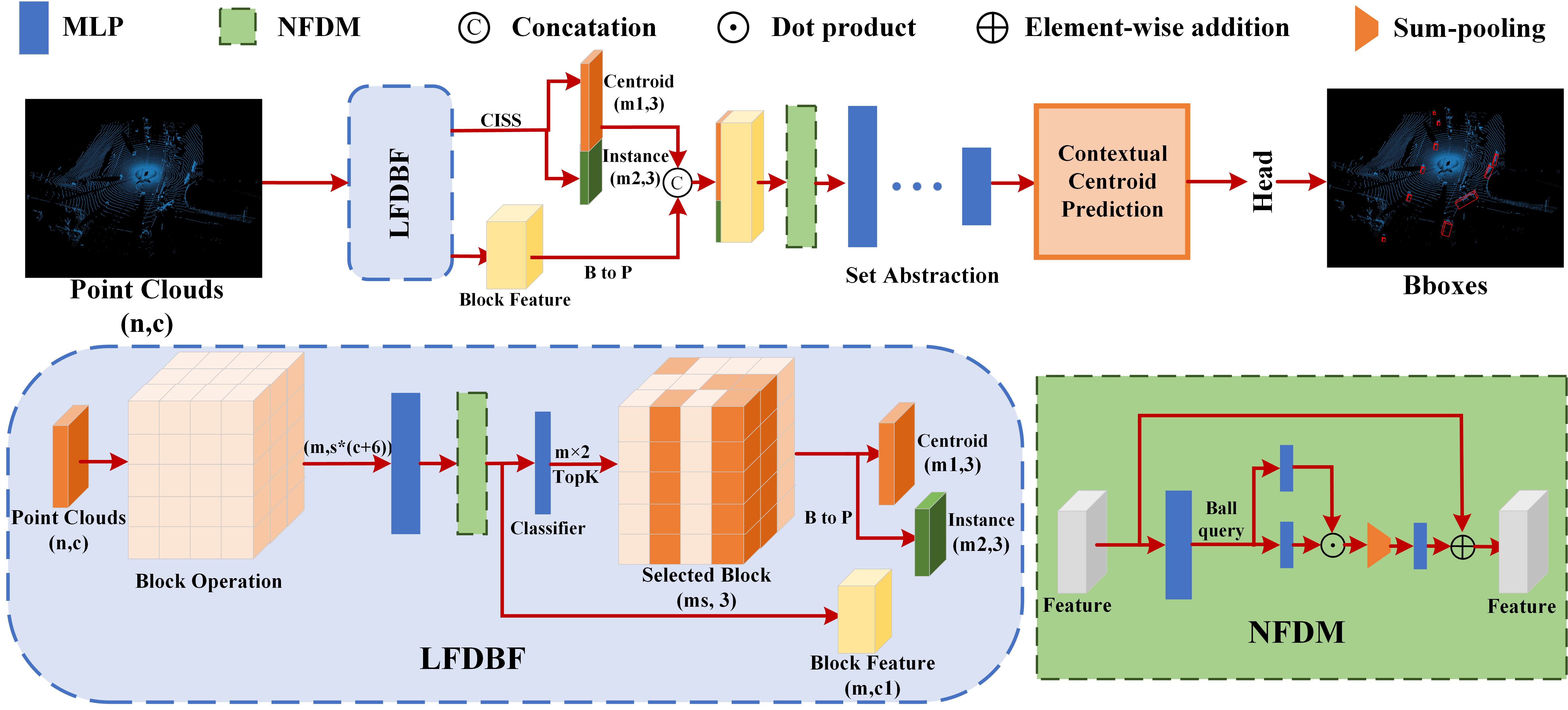}
    \caption{Diagram of IC-FPS framework. B to P means all points in the block are selected. $n$, $m$, $ms$, $m1$, and $m2$ represent the input number of point cloud, the number of valid blocks, the number of selected blocks, the number of selected centorid points and the number of selected Instance points, respectively. blocks colored in orange are selected as the foreground blocks. $c$ and $c1$ represent the number of input channel and feature channel respectively.}
    \label{fig: fig2}
\end{figure*}

As shown in Figure \ref{fig: fig2}, IC-FPS combines LFDBF and CISS to classify foreground and background regions, and conduct highly efficient downsampling of point cloud.
Then Neighborhood Feature Diffusion Module (NFDM) further enlarges diffusion range of foreground block features and reduces information loss due to downsampling. 
IC-FPS can be inserted to any point-based 3D object detection models and perform end-to-end training.

\subsection{Local Feature Diffusion based Background Point Filter}
\label{sec:LFDBF}
Local Feature Diffusion based Background Point Filter (LFDBF) is proposed to efficiently remove background blocks.
Given a set of points $P=\left \{ p_{i}\mid i=1,\cdots ,N \right \} \in R^{n\times c} $, $n$ represents the number of input points, and $c$ represents the number of channels.
As shown in Figure \ref{fig: fig2}, we partition the set of points $P$ into blocks and derive a matrix of size $[m,s,c]$, where $m$ is the number of effective blocks, $s$ is the number of points in block.
For efficient extraction of local features in each block, we employ the PointPillars \cite{lang2019pointpillars} method to acquire relative positional information within blocks. And the matrix size is folded to $[m, s, (c+6)]$, where the additional six dimensions include relative distances between points to center points and centroid positions in each block.
Valid classification of foreground and background blocks require features of neighbouring blocks to be included in each block, while MLPs merely extract independent block features.
Inspired by RANDLA-Net \cite{hu2020randla}, we construct a Neighborhood Feature Diffusion Module (NFDM) to substitute expensive 3D convolution and extract inter-block neighbouring information more efficiently.
NFDM uses multi-scale ball query as the alternative of KNN in RANDLA-Net to accelerate processing speed.

\begin{figure}[h!]
    \centering
    \includegraphics[scale=0.45]{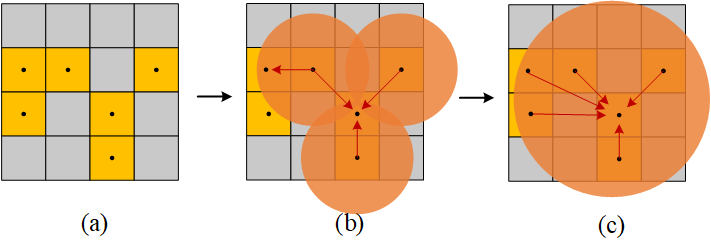}
    \caption{Diagram of Neighborhood Feature Diffusion Module. (a) represents block features, (b) represents feature diffusion range of valid block in the first NFDM, (c) represents feature diffusion range of valid block in the second NFDM.
    Yellow regions are valid blocks, orange regions represent diffusion range of features. Each block uses ball query to obtain its neighbourhood, and diffuses its features to other valid blocks in the neighbourhood.
    }
    \label{fig: fig3}
\end{figure}

As shown in Figure \ref{fig: fig3}, after neighbourhood range of each block is derived, we diffuse neighbourhood features of each block to other blocks in the neighbourhood. Hence, every block contains contextual information in the vicinity, which enables the network to rapidly classify foreground regions and reduces information loss of foreground point due to downsampling.

Features derived after NFDM are used to evaluate each block. MLP is employed as a classification network to compute confidence of blocks. The higher confidence, the more likely that the block is a "foreground block".
Blocks with confidence higher than threshold $\alpha$ are reckoned as the foreground block, whose features are represented as $F_{i}\in R^{\left ( ms\times c1 \right ) } $, where $c1$ is the number of channels,  $ms$ is the number of foreground blocks. All points in foreground blocks are defined as foreground points.

For better sampling sparse foreground regions at a distance in point cloud, we propose a Density-Distance Focal Loss $L_{DDFL}$ based on normal distribution and prevent distant instances from being filtered out. 
Density constraint $M_{Den}$ assigns various weights according to point density in the block.
\begin{equation}
\tag{1}
\label{eq:eq1}
M_{Den}=\frac{1}{\sqrt{2\pi \sigma }}exp\left ( -\frac{(\frac{N_{v} }{N_{Max} }-\mu)^{2}   }{2\alpha ^{2} }  \right )\times \sqrt{2\pi \sigma }
\end{equation}

where $\mu$ and $\sigma$ are position and scale parameters of normal distribution. 
$N_{v} $ and $N_{max}$ represent the number of valid points in block and the maximum value, respectively. 
Distance constraint $M_{Dis}$ assigns more weights on objects at a distance, and is expressed as follows,
\begin{equation}
\tag{2}
\label{eq:eq2}
M_{Dis}=\frac{exp\left ( \frac{D}{M_{D} }  \right )}{e}
\end{equation}
where $D$ is the distance between the point and the origin in the coordinate system, $M_{D} $ is the distance from the farthest point to the origin. And Density-Distance focal loss $L_{DDFL}$ is written as follows,
\begin{equation}
\tag{3}
\label{eq:eq3}
L_{DDFL}=\left ( 1-p_{t} \right )^{\gamma }log\left ( p_{t}  \right ) \times M_{Den}\times M_{Dis}
\end{equation}
where $p_t$ represents the difference between predicted value and ground truth value, $\gamma$ is an adjustable factor.

\subsection{Centroid-Instance Sampling Strategy}
\label{sec:CISS}
CISS aims to sample center points with high efficiency, and replace FPS in the first SA layer. 
We believe that reasons for the success of FPS are twofold. One is that FPS adaptively selects center points according to point density, i.e. more center points are selected in regions of high density.
The other is that FPS samples from the raw point cloud and preserves geometric structural information,  which is beneficial to improve accuracy of the following detection box regression. 
Therefore we add part of foreground points to center points, in order to increase sample density of instance objects. A centroid point offset module is constructed to restore the raw geometric structure of point cloud.

\textbf{Raw Instance Points Sampling.}
After foreground blocks are derived, centroid positions of blocks are calculated and noted as $D_i \in R^{ms\times3}$.
We sort centroid points according to classification confidence from the highest to the lowest, and select the highest $m_{1}$ points. 
In addition, we select $m_{2}$ points that have shortest distances to the origin. 
Both instance points and centroid points are regarded as the center points for the first SA layer in subsequent models. Algorithm \ref{alg:1} is the detailed procedure of CISS.

\textbf{Block Centroid Points Offset.} 
Adding positional information of the original point cloud to the following network layer by layer helps adapting to the original point cloud structure.
However there exist deviations between centroid and actual point cloud position.
Direct use of centroids might cause losing the original positional information, and the model to fail in predicting accurate size of bounding box during regression.
Consequently, we propose a centroid point offset module that moves centroid to the nearest instance point, for effective restoration on original size of targets. And the centroid point offset loss function $L_{CB}$  is written as follows, 
\begin{equation}
\tag{4}
\label{eq:eq4}
L_{CB} =SmoothL1\left ( \left \lfloor b \right \rfloor ,  \tilde{\left \lfloor b \right \rfloor }  \right )
\end{equation}
where $b$ is the predicted offset between the centroid and its nearest instance point, $\tilde{b} $ represents the actual offset of the centroid point to the nearest instance point. $\left \lfloor  \right \rfloor $ denote whether centroid is present in the instance box or not.

\begin{algorithm}
	\renewcommand{\algorithmicrequire}{\textbf{Input:}}
	\renewcommand{\algorithmicensure}{\textbf{Output:}}
	\caption{CISS}
	\label{alg:1}
	\begin{algorithmic}[1]
		\REQUIRE foreground block index $i$, foreground block feature $F_{v}\in R^{m_{s}\times c_{1} }$, classification confidence $\sigma$, raw point cloud position $P\in R^{n\times 3} $, hashtable that maps block to point $HashMap()$.
		\ENSURE center point feature $F_{C}\in R^{\left ( m_{1} + m_{2} \right )\times \left (3 + c_{1}  \right )}  $.
		\STATE $m_{1}=$ the number of selected centroid points
		\STATE $m_{2}=$ the number of selected instance points 
		\STATE Calculate centroid point position of each foreground block $P_{ctr}\in R^{m_{s} \times 3} $ and centroid point feature $F_{ctr}\in R^{m_{s} \times (3+c1)}\gets  Concat(P_{ctr}, F_{v})$.
		\STATE Calculate foreground point position $P_{ins}\in R^{n_{s}\times3 } \gets HashMap(P, i)$ and foreground point feature $F_{ins} \in R^{n_{s} \times \left (  3+c_{1} \right ) }\gets Concat\left (P_{ins},HashMap\left (F_{v}, i\right )\right )$.
		\STATE Sort centroid points according to classification confidence $\sigma$ in descending order, and select the centroid feature  $F_{c} \in R^{m_{1} \times (3+c_{1})} $ from $F_{ctr}$.
		\STATE Select the instance feature $F_{i} \in R^{m_{2} \times (3+c_{1})} $ from $F_{ins}$ by distance.
        \STATE Derive center point feature by merging centroid feature and instance feature :
        $F_{C}\in  R^{\left ( m_{1}+m_{2}   \right )  \times \left ( 3+c_{1}  \right ) } \gets Concat(F_{c}, F_{i})$.
	\end{algorithmic}  
\end{algorithm}

\subsection{Centroid Feature Diffusion Sampling framework}
\label{sec:IC-FPS}
As a plug-and-play module, IC-FPS is compatible with any point-based 3D object detection models.
In Figure \ref{fig: fig2}, We apply another NFDM to enlarge diffusion range of each centroid feature. 
Through stacking two NFDMs, IC-FPS can effectively reduce information loss due to downsampling and replace the sophisticated first SA layer.\\
\textbf{Total Loss.} The proposed IC-FPS framework can be integrated with other models for end-to-end training.
Multiple loss functions are combined for optimization. Total loss includes Density-Distance focal loss $L_{DDFL}$, centroid point offset loss $L_{CB}$, classification loss $L_{cls}$ and bounding box generation loss $L_{box}$, as given in Equation \ref{eq:eq5}.
\begin{equation}
\tag{5}
\label{eq:eq5}
L_{total}=L_{DDFL}+L_{CB}+L_{cls}+L_{box}
\end{equation}

\begin{table*}[h!]
\centering 
\small
\begin{tabular}{c|c|cc|cc|cc|c}

\hline
\makebox[0.02\textwidth][c]{Method} & \makebox[0.01\textwidth][c]{Type} & \makebox[0.01\textwidth][c]{\makecell{Veh.(L1) \\ mAP/mAPH}} & \makebox[0.01\textwidth][c]{\makecell{Veh.(L2) \\ mAP/mAPH}} & \makebox[0.01\textwidth][c]{\makecell{Ped.(L1) \\ mAP/mAPH}} & \makebox[0.01\textwidth][c]{\makecell{Ped.(L2) \\ mAP/mAPH}} & \makebox[0.01\textwidth][c]{\makecell{Cyc.(L1) \\ mAP/mAPH}} & \makebox[0.01\textwidth][c]{\makecell{Cyc.(L2) \\ mAP/mAPH}} & \makebox[0.01\textwidth][c]{Speed} \\ \hline
SECOND \cite{yan2018second}      & 1-stage & 68.03/67.44 & 59.57/59.04 & 43.49/23.51 & 37.32/20.17 & 35.94/28.34 & 34.60/27.29  & \underline{28.6} \\ 
Pointpillars \cite{lang2019pointpillars} & 1-stage & 60.67/59.79 & 52.78/52.01 & 43.49/23.51 & 37.32/20.17 & 35.94/28.34 & 34.60/27.29 & 25.0 \\ 
CenterPoint \cite{yin2021centerpoint} & 1-stage & 71.33/70.76 & 63.16/62.65 & 72.09/\underline{65.49} & 64.27/\underline{58.23} & 68.68/\underline{67.39} & 66.11/\underline{64.87} & 18.8 \\ 
PV-RCNN \cite{shi2020pv} & 2-stage & \underline{74.06}/\underline{73.38} & \underline{64.99}/\underline{64.38} & 62.66/52.68 & 53.80/45.14 & 63.32/61.71 & 60.72/59.18 & 3.7  \\ 
$Part-A^{2}$ \cite{shi2020parta2} & 2-stage & 71.82/71.29 & 64.33/63.82 & 63.15/54.96 & 54.24/47.11 & 65.23/63.92 & 62.61/61.35 & 8.8  \\ \hline
3DSSD* \cite{yang20203dssd}      & 1-stage & 71.27/70.75 & 62.78/62.32 & 54.57/48.68 & 46.79/41.68 & 62.00/60.74 & 59.64/58.44 & 4.9  \\ 
\textbf{IC-FPS-M + 3DSSD} & \textbf{1-stage} & \textbf{71.72/71.23} & \textbf{63.16/62.71} & \textbf{56.57/49.66} & \textbf{48.51/42.52} & \textbf{65.03/63.76} & \textbf{62.56/61.34} & \textbf{6.7}  \\ \hline
SASA* \cite{chen2022sasa}      & 1-stage & 71.45/71.01 & 62.99/62.57 & 63.17/56.99 & 54.11/49.01 & 63.82/62.24 & 61.45/60.18 & 7.3  \\ 
\textbf{IC-FPS-S + SASA} & \textbf{1-stage} & \textbf{71.63/71.14} & \textbf{63.17/62.73} & \textbf{63.80/57.88} & \textbf{55.03/49.86} & \textbf{64.49/63.23} & \textbf{62.04/60.83} & \textbf{8.8}  \\ \hline
IA-SSD \cite{zhang2022not}      & 1-stage & 70.53/69.67 & 61.55/60.80 & 69.38/58.47 & 60.30/50.73 & 67.67/65.30 & 64.98/62.71 & 15.5  \\ 
\textbf{IC-FPS-S + IA-SSD} & \textbf{1-stage} & \textbf{70.88/69.91} & \textbf{61.75/60.96} & \textbf{73.39/64.06} & \textbf{64.27/56.01} & \textbf{68.20/66.47} & \textbf{65.79/64.02} & \textbf{25.2}  \\ 
\textbf{IC-FPS-L + IA-SSD} & \textbf{1-stage} & \textbf{71.47/70.81} & \textbf{62.84/62.25} & \textbf{\underline{73.80}/64.18} & \textbf{\underline{64.80}/56.09} & \textbf{\underline{68.71}/66.65} & \textbf{\underline{66.17}/64.20} & \textbf{19.2}   \\ \hline
\end{tabular}
\caption{Quantitative comparison experiments on the Waymo $val$ set for 3D object detection.  $20\%$ of training set $(~32K)$ is used for training. Evaluation metrics are mean average precision (mAP) and mAP weighted by heading accuracy (mAPH). L1 and L2 denote \emph{Level1} and \emph{Level2}.  
Bold texts are our results and the best results are underlined.
For fair comparison, speed are derived by reproducing methods under OpenPCDet framework \cite{openpcdet2020}. Performance indexes follow the same configurations given in IA-SSD \cite{zhang2022not}, $\ast$ indicates results after reproduction of official open-source code.}
\label{tab:tab1}
\end{table*}

\section{Experiments and Discussions}
\subsection{Implementation Details}
We choose 3DSSD \cite{yang20203dssd}/SASA \cite{chen2022sasa}/IA-SSD \cite{zhang2022not} as the baseline and construct our model.
Firstly we partition the input point cloud into block. Block size is set to $[0.075, 0.075, 1]$.
In LFDBF module, point clouds are expanded via methods analogous to PointPillars, then fed into three MLP layers of size $(16, 16, 32)$.
The diffusion radius of the first NFDM is set to $4.0$ and the maximum number of diffusion points is set to $16$.
Confidence threshold is set to $0.45$.
We set $\mu = 0.5$ and $\sigma = 0.5$ in the DDFL (Equation \ref{eq:eq3}).
Centroid offset module in CISS contains two MLP layers with size of $(16, 3)$.
For the second NFDM, the diffusion radius is set to $[0.2, 0.8]$ and the maximum number of diffusion points is set to $16$.

We configure three IC-FPS modules with different number of samples in our experiment, which are IC-FPS-S/IC-FPS-M/IC-FPS-L. 
The maximum number of sampled centroid points and instance points are set to $16384/2048$, $26000/4096$ and $30720/8197$, respectively.

The same training strategy and model structure in each baseline is adopted in our experiment, except that the first SA layer is replaced by IC-FPS.
In IA-SSD experiments, batch size is set to 8, learning rate is set to 0.01, Adam\cite{kingma2014adam} optimizer is employed with weight decay set to $0.01$.
In 3DSSD and SASA experiment, batch size is set to 2, learning rate is set to 0.002.
All experiments are conducted on NVIDIA A40 GPU and AMD EPYC 7402 CPU.

\subsection{State-of-the-art Comparison}
\noindent \textbf{3D Detection on Waymo.}
To validate performance of IC-FPS on large-scale point cloud scenes, we conduct quantitative experiments on Waymo \cite{sun2020scalability} dataset.
Waymo dataset \cite{sun2020scalability} contains 160K samples in training set and 40K samples in validation set with two difficulty levels of challenge, $Level1$ and $Level2$.
We configure three IC-FPS with different number of samples, which are IC-FPS-S/IC-FPS-M/IC-FPS-L. 
And the maximum number of sampled centroids and instance point are set to (16384, 2048), (26000, 4096) and (30720, 8197).
In the last three SA layers, the number of samples are set to 4096, 2048 and 1024.
IoU threshold is 0.25.

\begin{table}[h!] 
\small
\centering 
\begin{tabular}{c|cc} \hline
Method & Para. (M) & \emph{FPS} \\ \hline
3DSSD \cite{yang20203dssd} & 4.79 & 2.5\\
\textbf{IC-FPS-M + 3DSSD} & $\bm{4.84}$ & $\bm{3.8}$\\
\hline
SASA \cite{chen2022sasa} & 4.83 & 2.6\\
\textbf{IC-FPS-S + SASA} & $\bm{4.88}$ & $\bm{4.0}$\\
\hline
IA-SSD \cite{zhang2022not} & 2.70 & 2.7\\
\textbf{IC-FPS-S + IA-SSD} & $\bm{2.74}$ & $\bm{12.9}$\\
\textbf{IC-FPS-L + IA-SSD} & $\bm{2.74}$ & $\bm{8.5}$ \\
\hline
\end{tabular}
\caption{Inference Speed Comparison experiment of Waymo 3D Object Detection, \emph{FPS} indicates that the inference speed of the model when batchsize=1, Para. indicates that the parameter size.} 
\label{tab:tab2}
\end{table}

\begin{table*}[h!]
\small
\centering 

\begin{tabular}{c|cc|cc|cc|c} \hline 
\makebox[0.02\textwidth][c]{Method} & \makebox[0.01\textwidth][c]{\makecell{Veh.(L1) \\ mAP/mAPH}} & \makebox[0.01\textwidth][c]{\makecell{Veh.(L2) \\ mAP/mAPH}} & \makebox[0.01\textwidth][c]{\makecell{Ped.(L1) \\ mAP/mAPH}} & \makebox[0.01\textwidth][c]{\makecell{Ped.(L2) \\ mAP/mAPH}} & \makebox[0.01\textwidth][c]{\makecell{Cyc.(L1) \\ mAP/mAPH}} & \makebox[0.01\textwidth][c]{\makecell{Cyc.(L2) \\ mAP/mAPH}} & \makebox[0.01\textwidth][c]{\emph{FPS}} \\ \hline
IA-SSD \cite{zhang2022not}   & 70.53/69.67 & 61.55/60.80 & 69.38/58.47 & 60.30/50.73 & 67.67/65.30 & 64.98/62.71 & 2.7  \\ 
IA-SSD$\dagger $    \cite{zhang2022not}  & 70.38/69.82 & 61.33/60.19 & 68.23/58.44 & 60.11/50.33 & 66.84/65.06 & 64.32/62.61 & 8.3 \\ 
\textbf{IC-FPS-L}   & \textbf{\underline{71.47}/\underline{70.81}} & \textbf{\underline{62.84}/\underline{62.25}} & \textbf{\underline{73.80}/\underline{64.18}} & \textbf{\underline{64.80}/\underline{56.09}} & \textbf{\underline{68.71}/\underline{66.65}} & \textbf{\underline{66.17}/\underline{64.20}} & \textbf{8.5} \\
\textbf{IC-FPS-L$\dagger$}   & \textbf{71.19/70.48} & \textbf{62.62/61.99} & \textbf{72.93/63.11} & \textbf{63.72/55.21} & \textbf{67.22/65.49} & \textbf{64.75/63.09} & \textbf{\underline{12.6}}  \\ \hline
\end{tabular}
\caption{Quantitative comparsion experiments on Waymo $val$ set for 3D object detection.
$\dagger$ indicates that the whole scene is divided into four parts to accelerate SA layer. Bold texts are our results and the best results are underlined.
\label{tab:tab3}
}
\end{table*}

\begin{figure}
    \centering
    \includegraphics[width=0.48\textwidth, height=0.23\textheight]{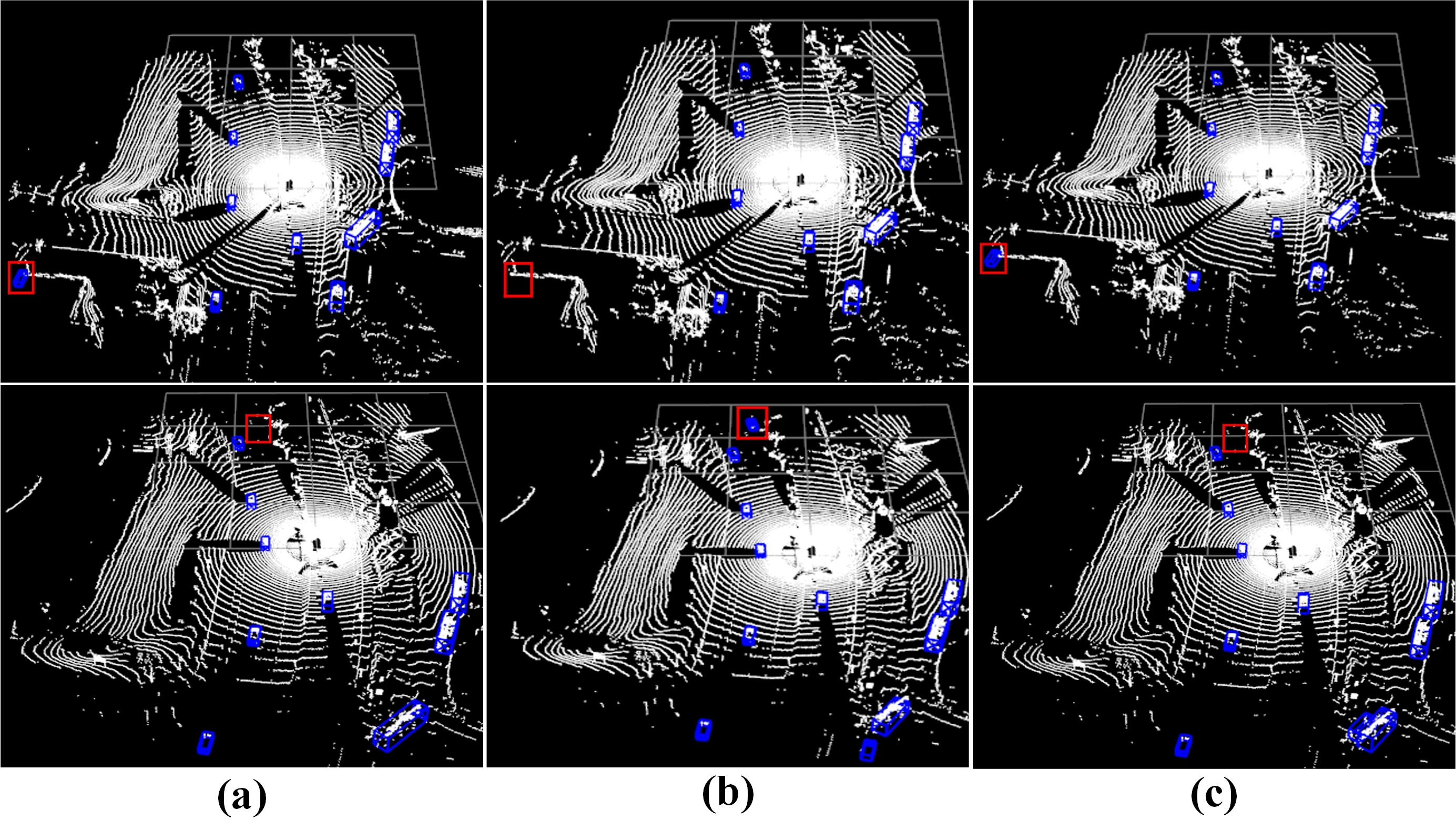}
    \caption{Visulization results of Waymo dataset $Vehicle$. (a) ground-truth, (b) IA-SSD, (c) IC-FPS+IA-SSD. Red boxes represent differences of various results.}
    \label{fig: fig4}
\end{figure}

\begin{figure}
    \centering
    \includegraphics[width=0.45\textwidth, height=0.33\textheight]{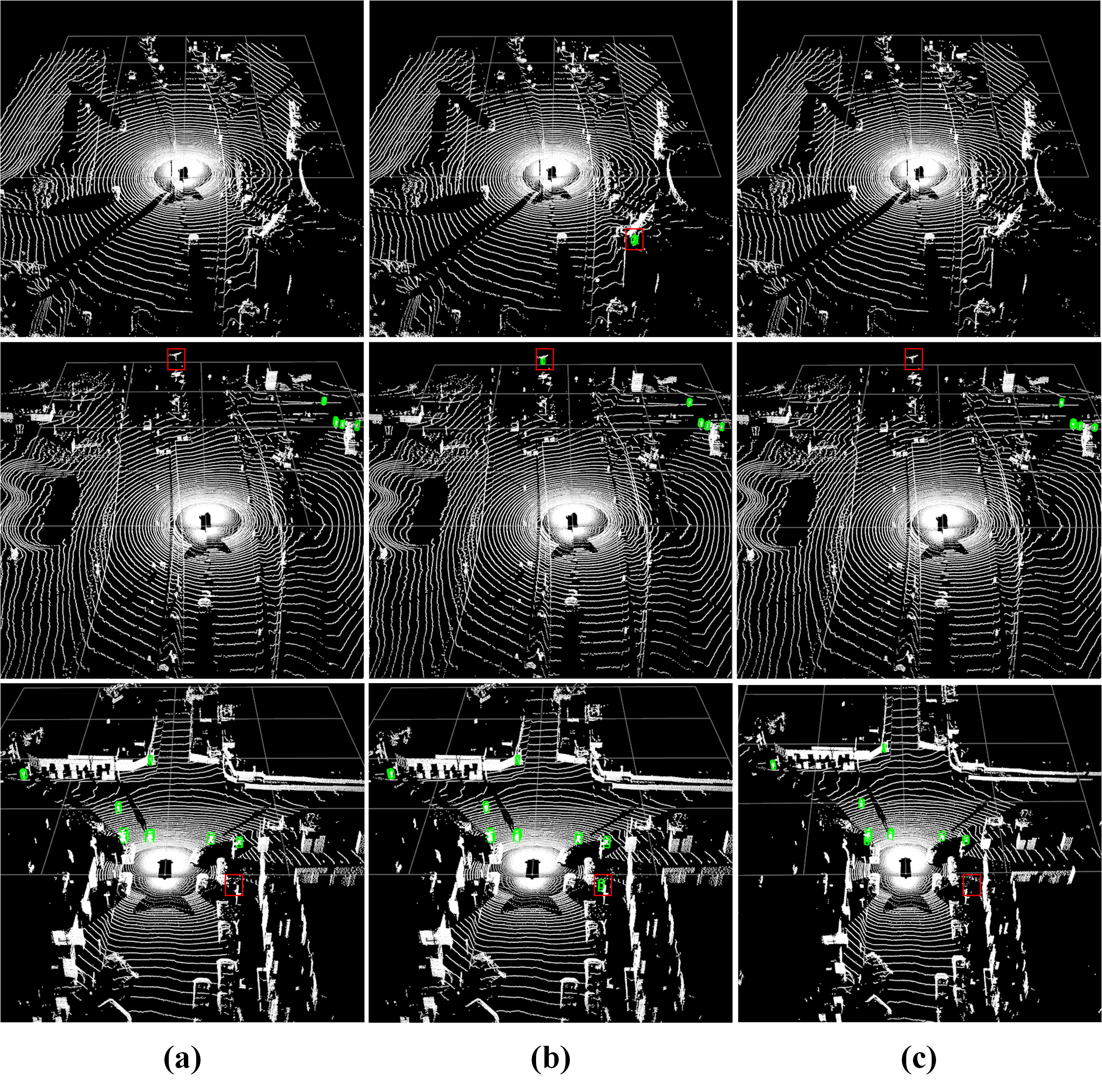}
    \caption{Complementary visualization of Waymo Dataset Pedestrian, (a) ground truth, (b) IA-SSD, (c) IC-FPS + IA-SSD. Bounding boxes in red are the difference, green are the Pedestrian.} 
    \label{fig: fig5}
\end{figure}

\begin{figure}
    \centering
    \includegraphics[width=0.45\textwidth, height=0.34\textheight]{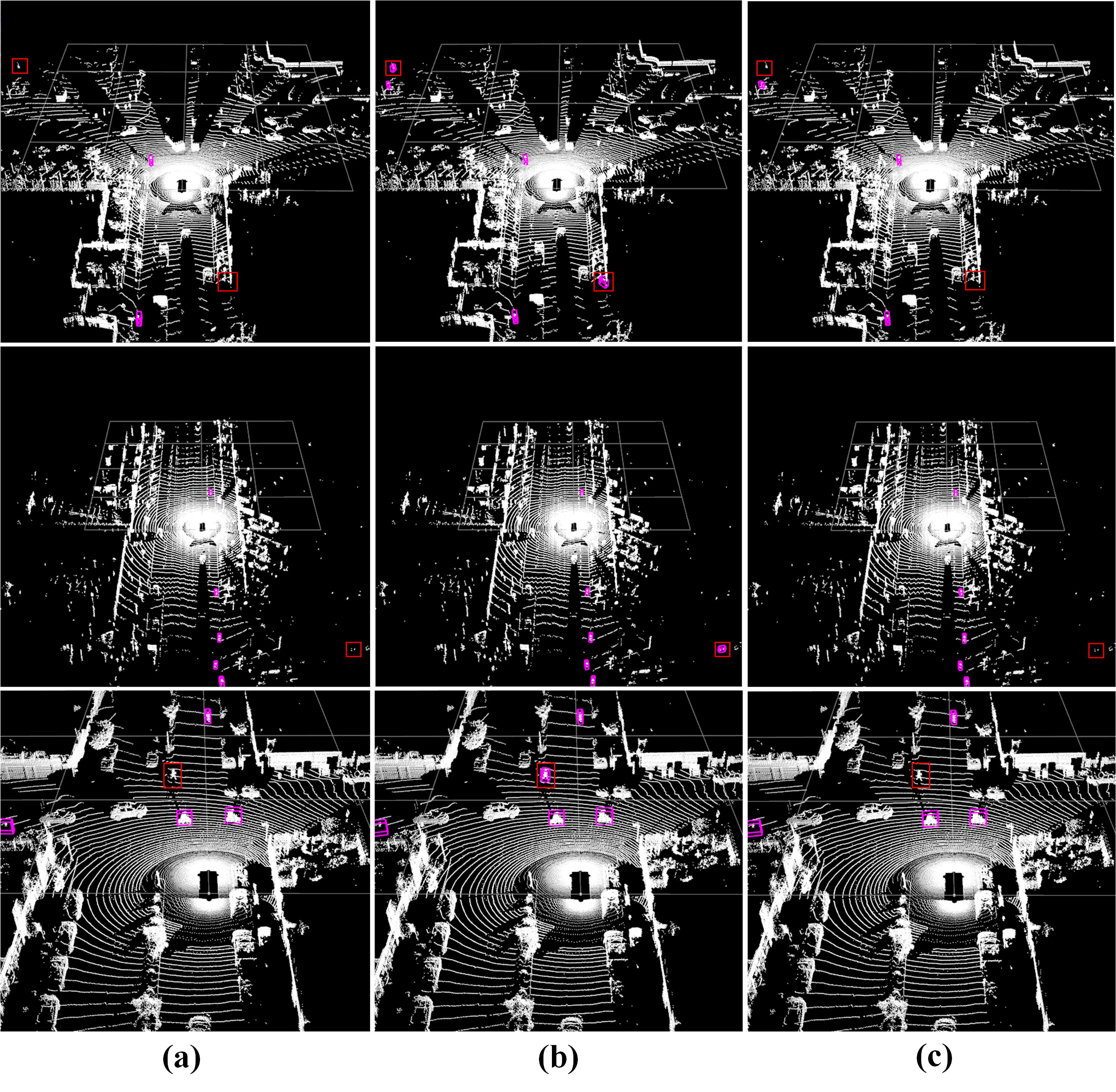}
    \caption{Complementary visualization of Waymo Dataset Cyclist, (a) ground truth, (b) IA-SSD, (c) IC-FPS + IA-SSD. Bounding boxes in red are the difference, purple are the cyclist. } 
    \label{fig: fig6}
\end{figure}

Table \ref{tab:tab1} demonstrates that the proposed IC-FPS significantly improves the inference speed and accuracy of baseline model.
Compared to the baseline model IA-SSD \cite{zhang2022not}, on \emph{Level1} IC-FPS(16384/2048) improves $0.35\%/0.24\%, 4.01\%/5.59\%, 0.53\%/1.17\%$ in terms of mAP/mAPH. Throughput is increased by 0.6 times.
For IC-FPS(30720+8192),
on \emph{Level2} mAP/mAPH are improved by $(1.14\%/1.45\%, 4.42\%/5.36\%, 1.04\%/1.49\%)$. Inference speed is improved by 20\%.
IC-FPS also effectively boosts accuracy and speed of 3DSSD and SASA.
IC-FPS(16384+2048) improves mAP/mAPH by $(10.12\%/8.95\%, 40.55\%/35.84\%, 38.13\%/36.73\%)$ on \emph{Level2}, compared to PointPillars that has comparable throughput. 
Visual results can be found in Figure \ref{fig: fig4}, \ref{fig: fig5} and \ref{fig: fig6}.

Table \ref{tab:tab2} reports inference speed of each baseline before and after IC-FPS.
Since CISS avoids heavy computational burden of the first SA layer, IA-SSD inference speed is increased by 3.7/2.1 times after inserting IC-FPS. And it becomes the first point-based 3D object detection model that realizes real-time detection in large-scale point cloud dataset.
As for 3DSSD and SASA, acceleration effect of IC-FPS is degraded because of multi-scale neighbourhood feature aggregation in the second SA layer. 

We adopt the FPS acceleration method in IA-SSD\cite{zhang2022not}, and partition the whole scene into four parts in Table \ref{tab:tab3}.
Although this method can increase inference speed, it degrades the model performance to some extent.
While after inserting IC-FPS, both model inference speed and model accuracy have obvious improvement.

\noindent \textbf{3D Detection on ONCE.}
ONCE (One millionN cSenEs) dataset \cite{mao2021once} includes 1 million 3D scenes and 7 million corresponding 2D images. Recording duration of 3D scenes lasts for 144 driving hours. And it covers various weathers, traffic conditions, time and zones. 
\begin{table*}[h!]
\centering 
\small
\begin{tabular}{c|c|c|c|c}

\hline
\makebox[0.02\textwidth][c]{Method}  &  \makebox[0.21\textwidth][c]{\makecell{Vehicle \\ Overall\ \ 0-30m\ 30-50m\ $>$50m}} &  \makebox[0.21\textwidth][c]{\makecell{Pedestrian \\ Overall\ \ 0-30m\ 30-50m\ $>$50m}} & \makebox[0.21\textwidth][c]{\makecell{Cyclist \\ Overall\ \ 0-30m\ 30-50m\ $>$50m}} & \makebox[0.01\textwidth][c]{mAP} \\ \hline
PointPillars \cite{yan2018second} & 68.57\quad80.86\quad62.07\quad47.04 & 17.63\quad19.74\quad15.15\quad10.23 & 46.81\quad58.33\quad40.32\quad25.86 & 44.3  \\ 
SECOND \cite{lang2019pointpillars} & 71.19\quad84.04\quad63.02\quad47.25 & 26.44\quad29.33\quad24.05\quad18.05 & 58.04\quad69.96\quad52.43\quad34.61 & 51.9 \\ 
PV-RCNN \cite{shi2020pv} & \underline{77.77}\quad\underline{89.39}\quad\underline{72.55}\quad\underline{58.64} & 23.50\quad25.61\quad22.84\quad17.27 & 59.37\quad71.66\quad52.58\quad36.17  & 53.6   \\ 
PointRCNN \cite{shi2019pointrcnn} & 52.09\quad74.45\quad40.89\quad16.81 & 4.28\quad\ \ 6.17\quad\ \ 2.40\quad\ \ 0.91 & 29.84\quad46.03\quad20.94\ \quad5.46 & 28.7 \\ 
\hline
IA-SSD \cite{zhang2022not} & 70.30\quad83.01\quad62.84\quad47.01&39.82\quad47.45\quad\underline{32.75}\quad18.99 & 62.17\quad73.78\quad56.31\quad\underline{39.53} & 57.4 \\ 
\textbf{+IC-FPS-L} & \textbf{70.56}\quad\textbf{82.73}\quad\textbf{64.47}\quad\textbf{48.75}&\textbf{\underline{40.09}}\quad\textbf{\underline{47.64}}\quad\textbf{32.57}\quad\textbf{\underline{20.51}} & \textbf{\underline{62.80}}\quad\textbf{\underline{75.64}}\quad\textbf{\underline{57.65}}\quad\textbf{38.14} & \textbf{\underline{57.8}} \\ 
\hline
\end{tabular}
\caption{Quantitative comparison experiments on ONCE validation set. Bold texts are our results, underlined texts are the best results. Performance indexes follow the same configurations given in IA-SSD \cite{zhang2022not}.}
\label{tab:tab4}
\end{table*}

We report our results on ONCE dataset in Table \ref{tab:tab4}. IC-FPS outperforms the baseline in all three categories.

\begin{figure}
    \centering
    \includegraphics[width=0.475\textwidth, height=0.3\textheight]{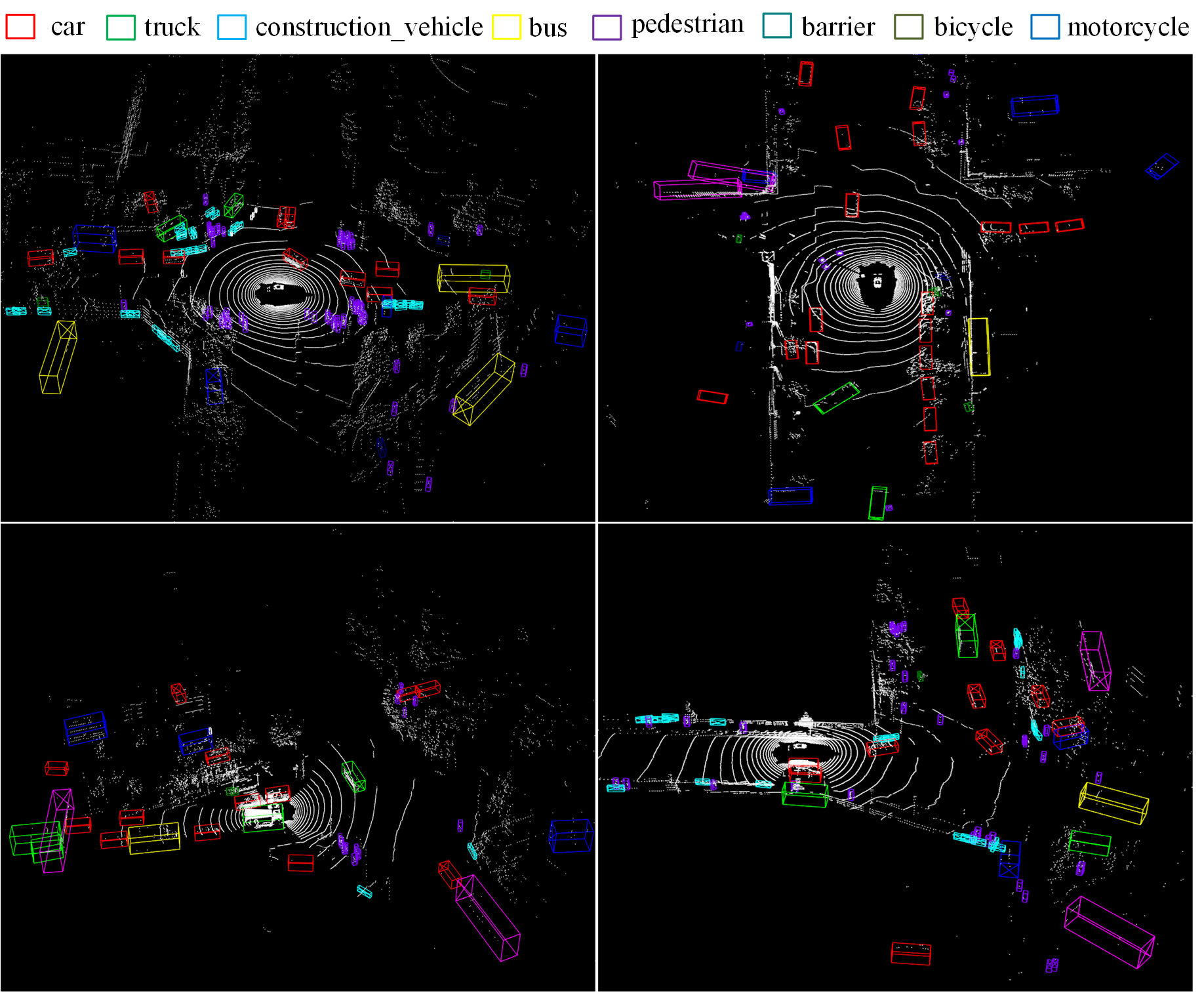}
    \caption{Complementary visualization of our methods on nuScenes Dataset.
    } 
    \label{fig: fig7}
\end{figure}

\noindent \textbf{3D Detection on nuScenes.}
We further conduct comparison experiments on nuScenes dataset \cite{caesar2020nuScenes} to verify the robustness of IC-FPS.
nuScenes dataset contains 40K annotated keyframes with 23 object categories.
mAP and NDS denote mean Average Precision and nuScenes detection score.
Visualization results are shown in Figure \ref{fig: fig7}.

\begin{table*}[h!]
\small
\centering 

\begin{tabular}{c|cc|cccccccccc|c} \hline 
\makebox[0.02\textwidth][c]{Method} & \makebox[0.02\textwidth][c]{NDS} & \makebox[0.02\textwidth][c]{mAP} & \makebox[0.02\textwidth][c]{Car} & \makebox[0.02\textwidth][c]{Truck} & \makebox[0.02\textwidth][c]{Bus} & \makebox[0.02\textwidth][c]{Tra.} & \makebox[0.02\textwidth][c]{C.V.} & \makebox[0.02\textwidth][c]{Ped.} & \makebox[0.02\textwidth][c]{Motor} & \makebox[0.02\textwidth][c]{Bicy.} & \makebox[0.02\textwidth][c]{T.C.} & \makebox[0.02\textwidth][c]{Barrier} & \makebox[0.02\textwidth][c]{\emph{FPS}} \\ \hline
3D-CVF   \cite{yoo20203dcvf} & 49.8 & 42.2 & 79.7 & 37.9 & 55.0 & 36.3 & -   & \underline{71.3} & 37.2 & -   & \underline{40.8} & 47.1 & -    \\ 
SASA    \cite{chen2022sasa} & \underline{61.0} & \underline{45.0} & 76.8 & 45.0 & 66.2 & \underline{36.5} & 16.1 & 69.1 & 39.6 & \underline{16.9} & 29.9 & \underline{53.6} & 1.9 \\ 
3DSSD    \cite{yang20203dssd}      & 56.4 & 42.6 & \underline{81.2} & \underline{47.2} & 61.4 & 30.5 & 12.6 & 70.2 & 36.0 & 8.6  & 31.1 & 47.9 & 2.0  \\ \hline
IA-SSD   \cite{zhang2022not}      & 48.8 & 44.0 & 73.8 & 45.1 & \underline{67.0} & 29.7 & \underline{17.0} & 66.9 & 40.6 & 14.6 & 32.0 & 53.2 & 2.4  \\ 
\textbf{IC-FPS + IA-SSD} & \textbf{49.1} & \textbf{44.2} & \textbf{74.8} & \textbf{45.7} & \textbf{66.2} & \textbf{29.8} & \textbf{16.8} & \textbf{70.5} & \textbf{\underline{40.7}} & \textbf{14.8} & \textbf{29.4} & \textbf{\underline{53.6}} & \textbf{\underline{6.8}} \\ \hline
\end{tabular}
\caption{Comparison results of various methods on nuScenes $val$ set.
Bold texts are our results and best results are underlined.
For fair comparison, inference time and performance evaluation metrics are derived by reproducing methods under OpenPCDet \cite{openpcdet2020}.
}
\label{tab:tab5}
\end{table*}
We report our results on nuScenes dataset in Table \ref{tab:tab5}.
Our IC-FPS outperforms baseline model by $0.3\%$ and $0.2\%$ in NDS and mAP while accelerating inference by 1.8 times.

\begin{table*}[h!] 
\small
\centering 

\begin{tabular}{ccc|ccc} \hline
NFDM & DDFL & Ctr-bias & \makecell{Veh.(L1) \\mAP} & \makecell{ped.(L1) \\ mAP} & \makecell{Cyc.(L1) \\ mAP}       \\ \hline
$\times $ & $\times $ & $\times $ & 64.89 & 59.71 & 61.49  \\
$\surd $  & $\times $ & $\times $ & 67.02 & 60.75 & 63.66  \\
$\surd $  & $\surd $  & $\times $ & 68.15 & 68.33 & 67.18  \\
\textbf{$\surd $ } & \textbf{$\surd $}  & \textbf{$\surd $}  & \textbf{71.47} & \textbf{73.80} & \textbf{68.71}  \\
\multicolumn{3}{c}{\textbf{Improvement}} & \textbf{+6.58} & \textbf{+14.09} & \textbf{+7.22}  \\ \hline
\end{tabular}
\caption{Ablation study on IC-FPS. We report mAP value of Waymo\cite{sun2020scalability} dataset on $LEVEL1$, where NFDM stands for Neighbourhood Feature Diffusion Module, DDFL denotes Density-Distance Focal Loss, and Ctr-bias represents Centroid Point Offset Module.}
\label{tab:tab6}
\end{table*}

\begin{table*}[h!] 
\small
\centering 
\begin{tabular}{ccc|ccc|c} \hline
Ins. Points & Ctr. Points & FPS & \makecell{Veh. (L1) \\ mAP} & \makecell{ped. (L1) \\ mAP} & \makecell{Cyc. (L1) \\ mAP}  & \emph{FPS} \\ \hline
$\times $ & $\times $ & $\surd $ & 70.53 & 69.38 & 67.67 & 2.7 \\ 
$\times $  & $\surd $ & $\times $ & 70.17 & 70.51 & 67.08 & 13.3 \\ 
\textbf{$\surd $}  & \textbf{$\surd $ } & \textbf{$\times $} & \textbf{70.88} & \textbf{73.39} & \textbf{68.20} & \textbf{12.9} \\
\multicolumn{3}{c}{\textbf{Improvement}}& \textbf{+0.35} & \textbf{+4.01} & \textbf{+0.53} & \textbf{+10.2} \\ \hline
\end{tabular}
\caption{Ablation study of IC-FPS variants using different downsampling strategies in the first layer. mAP values of Waymo\cite{sun2020scalability} dataset in L1 difficulty are reported. Ins. Points represent using instance points as center points. Ctr. Points represent using centroid points as center points. FPS represents using the farthest distance sampling strategy to calculate center points. \textit{FPS} represents the model inference speed when batch size is set to 1. }
\label{tab:tab7}
\end{table*}

\subsection{Ablation Study}
\begin{figure}
    \centering
    \includegraphics[width=0.48\textwidth, height=0.12\textheight]{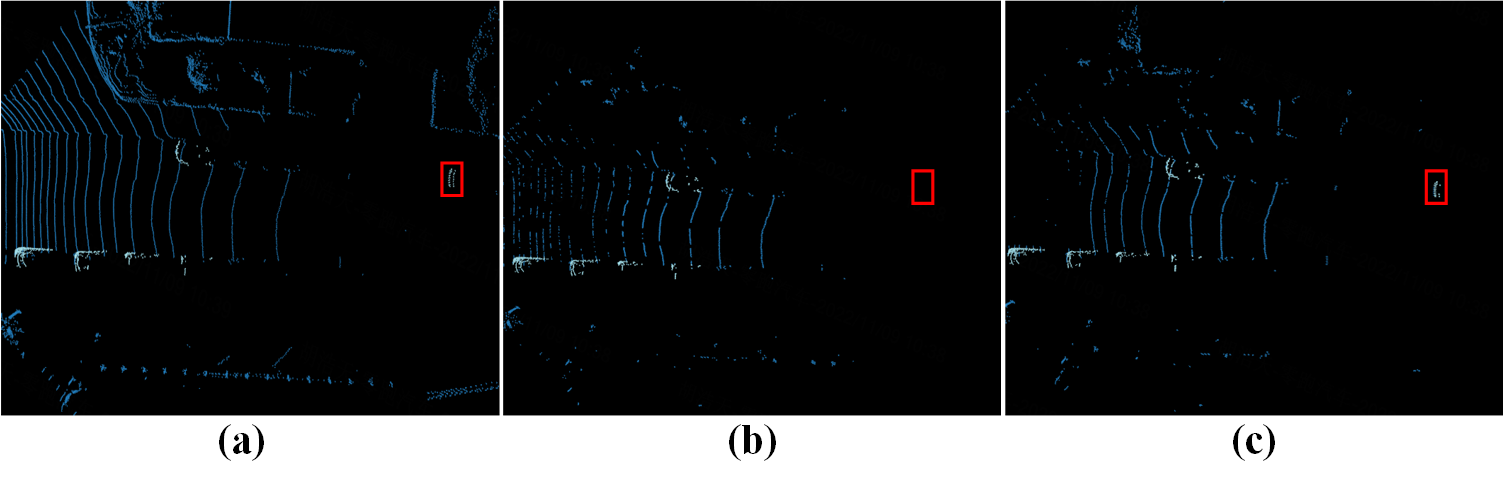}
    \caption{Visualisation results before and after DDFL. (a) original point cloud input, (b) Without DDFL, (c) With DDFL. White dots are foreground points. Bounding boxes in red are the difference.}
    \label{fig: fig8}
\end{figure}

We conduct various ablation experiments on Waymo dataset \cite{sun2020scalability} .
As shown in Table \ref{tab:tab6}, we construct three different IC-FPS variants by ablating each proposed module.
Experiment results indicate that:
(1) NFDM alleviates the information loss caused by downsampling while enhancing model performance.
(2) LFDBF can separate foreground and background points better by 
adding distance and density constraints, especially for small objects in the far distance.
(3) Retrieving raw point cloud geometry from blocks helps subsequent network to better capture scale information of instance object.
Figure \ref{fig: fig8} compares results before and after adding DDFL. We can find that distant targets are sampled, which effectively improves model recall rate.

\begin{figure}
    \centering
    \includegraphics[width=0.48\textwidth, height=0.21\textheight]{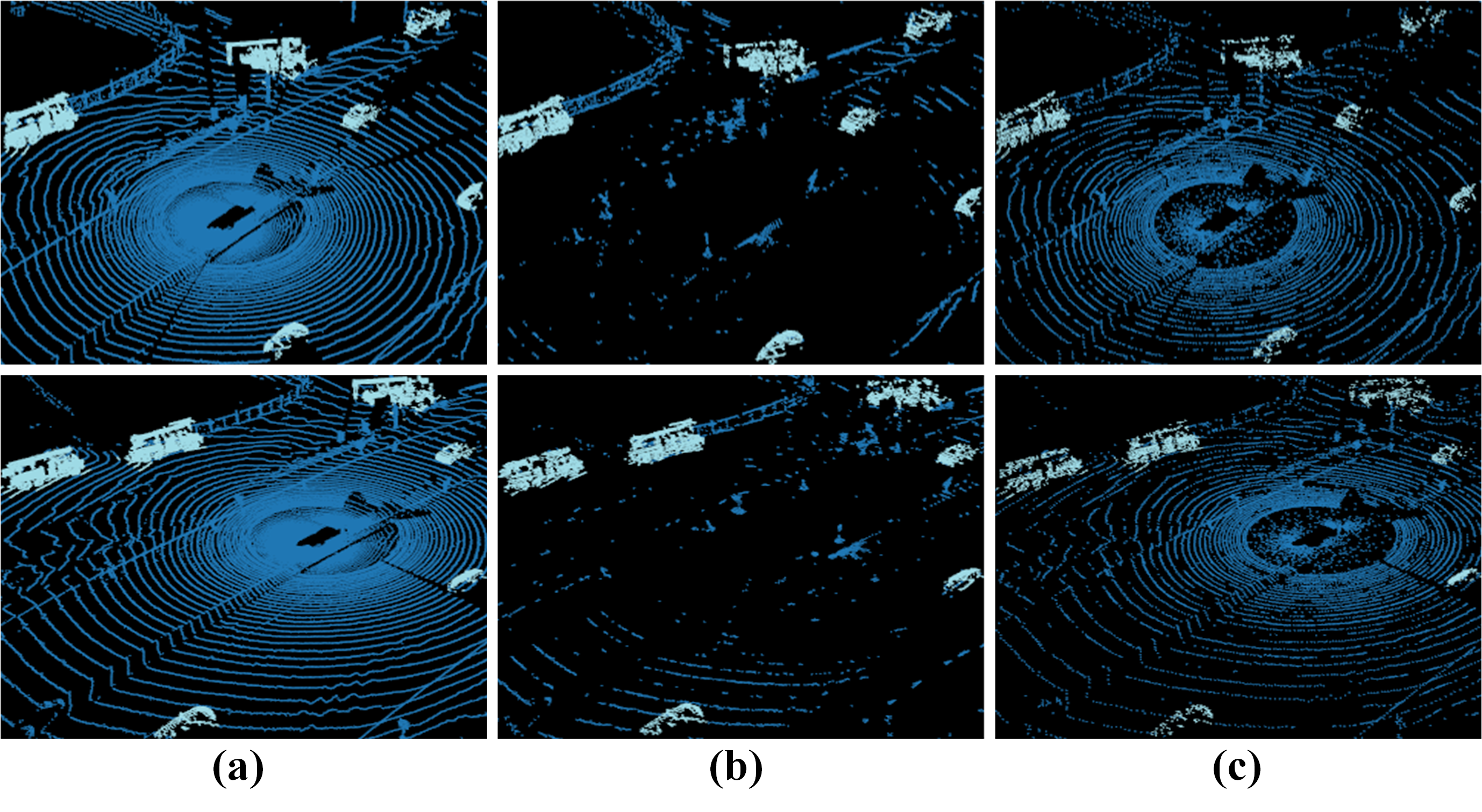}
    \caption{Distribution of sample points by various sampling strategies, (a) original point cloud input, (b) CISS, (c) FPS. White dots are real instance points, blue dots are background points.}
    \label{fig: fig9}
\end{figure}

Table \ref{tab:tab7} reports ablation study of IC-FPS on various downsampling strategies. 
It is shown that Centroid-based downsampling significantly improves the inference speed of IC-FPS. But it does not refer to the density distribution of instance target, resulting in decreasing accuracy.
Combining instance points with centroid points as sampling center not only improves the performance of baseline by $0.35\%$, $4.01\%$ and $0.53\%$, but also still realizes real-time detection (12.9 \emph{FPS}).
As given in Figure \ref{fig: fig9}, compared to FPS, CISS samples more foreground points while excluding most of the background points, which allows the model to better focus on the instance target.

\section{Conclusions}
\vspace{-0.5em}
In this paper, we propose an efficient IC-FPS module for accelerating point-based 3D object detection model
It contains a fast center sampling strategy CISS, which effectively avoids huge computational burdens brought by FPS strategy in the first SA layer.
To extract sufficient information from foreground points and exclude invalid background points, LFDBF is built to reduce information loss during downsampling.
Experiment results on Waymo and nuScenes datasets have demonstrated that our proposed IC-FPS module is capable of improving baseline model performance and increasing inference speed significantly. 
Moreover, it is the first time that real-time detection of point-based model is realized in large-scale point cloud scenes. 
And it indicates a promising future for point-based 3D object detection methods to be applied in autonomous driving and other related areas.

{\small
\bibliographystyle{ieee_fullname}
\bibliography{iccv}
}
\end{document}